\documentclass[11pt,a4paper]{article}
\usepackage[hyperref]{acl2019}
\usepackage{times}
\usepackage{latexsym}
\usepackage{url}
\usepackage{booktabs}
\usepackage{adjustbox}
\usepackage{xcolor}
\usepackage{multirow}
\usepackage{amsmath,amssymb}

\usepackage{algorithm}
\usepackage{algorithmic}

\aclfinalcopy 

\title{Understanding Learning Dynamics for Neural Machine Translation}

\author{{Conghui Zhu\textsuperscript{$\epsilon$}, Guanlin Li\textsuperscript{$\epsilon$}}~\thanks{~~~Preprint, work done at Tencent AI Lab.}~~, Lemao Liu\textsuperscript{$\lambda$}
, Tiejun Zhao\textsuperscript{$\epsilon$}, Shuming Shi\textsuperscript{$\lambda$} \\
\textsuperscript{$\epsilon$}Harbin Institute of Technology, \textsuperscript{$\lambda$} Tencent AI Lab \\
    \{chzhu, tjzhao\}@hit.edu.cn, \{epsilonlee.green\}@gmail.com,\\
    \{redmondliu, shumingshi\}@tencent.com \\
}

\date{}

\begin{document}
\maketitle
\begin{abstract}
Despite the great success of NMT, there still remains a severe challenge: it is hard to interpret the internal dynamics during its training process.  In this paper we propose to understand learning dynamics of NMT by using a recent proposed technique named Loss Change Allocation (LCA)~\citep{lan-2019-loss-change-allocation}. As LCA requires calculating the gradient on an entire dataset for each update, we instead present an approximate to put it into practice in NMT scenario. 
Our simulated experiment shows that such approximate calculation is efficient and is empirically proved to deliver consistent results to the brute-force implementation. In particular, extensive experiments on two standard translation benchmark datasets reveal some valuable findings.
\end{abstract}

\section{Introduction}
Neural Machine Translation (NMT) has witnessed a great success in recent years~\cite{Wu2016GooglesNM,gehring2017convolutional,vaswani2017attention}.
The main reason of its success is that it employs a mass of parameters to model
sufficient context for translation decision, and in particular enjoys an end-to-end flavor for training all these parameters.
Despite its success, there still remains a severe interpretation challenge for NMT: it is hard to understand its learning dynamics, i.e., 
{\em how do the trainable parameters affect a NMT model during its learning process}?

Understanding learning dynamics of neural networks is beneficial to identify the potential training issues and further improve training protocols for neural networks~\citep{Smith2017DontDT,McCandlish2018AnEM}.
Existing works on understanding learning dynamics have been extensively investigated in classification tasks~\cite{shwartz2017opening,NIPS2017_7188,li2018visualizing,bottou2018optimization,Combes2018OnTL}.
Unlike neural networks for classification tasks, NMT involves in a complex architecture with massive parameters and requires large scale data for training, which makes it more difficult to understand its learning dynamics.
To the best of our knowledge, there are no attempts at understanding the mechanism of learning dynamics for NMT, 
although it is acknowledged that the training process is critical to make advanced NMT architectures successful.

In this paper, we thereby propose to understand learning dynamics of NMT.
Specifically, we use a technique named Loss Change Allocation (LCA) to decompose the overall loss according to individual parameter for each update during the training process as the moment LCA value~\cite{lan-2019-loss-change-allocation}. By summing up the LCA values of a parameter between two update time steps, we are able to quantify how effective certain groups of parameter are to the loss decrease in this learning phase.
We utilize LCA to analyze the learning dynamics of model parameters for model's fitting ability on the training data. Since the original LCA requires calculating gradient on the entire training data for each update, whose brute-force implementation is impractical for standard translation tasks,
we instead approximately calculate it on a stochastic mini-batch from training or test data for speedup.
Our simulation shows that such approximate calculation is efficient and is empirically proved to deliver consistent results compared to the brute-force implementation. Furthermore, extensive experiments on two standard translation tasks reveal the following findings:
\begin{itemize}
\item Parameters of the encoder (decoder) word embeddings and the softmax matrix contribute very little to loss decrease;
\item Parameters of both the last layer in encoder and decoder contribute most to loss decrease than other layers;
\item Word embeddings for frequent words contribute far more to the loss decrease than those for infrequent words.
\end{itemize}

\section{Methods}


\subsection{Loss Change Allocation}

Loss Change Allocation (LCA)
functions as a microscope for investigating deeply into the
training process of any models trained with
stochastic gradient methods~\citep{lan-2019-loss-change-allocation}.
It is an optimizer-agnostic methods for probing into fine-grained learning dynamics.
In raw wordings, LCA tracks the contribution of each parameter
$\theta^i \in \theta$ at each gradient update of
the loss change during the training
process, where $i \in [K]$ and $K$ is
the number of model parameters.
The basic idea of LCA is to take advantage of the
first order Taylor expansion to approximate
the loss change at each mini-batch update.

Recall that at each update step $t$,
the optimizer (e.g. SGD) samples a mini-batch $B_t$ from
the training data for forward computation
and then backwards to update the parameters
from $\theta_t$ to $\theta_{t+1}$.
Given a dataset $\mathcal{D}$, formally, 
the moment loss change over all the model parameters are
approximated and decomposed on LCA of each parameter $\theta_i$
as follows:
\begin{equation}
\begin{split}
    & \mathcal{L}(\theta_{t+1}; \mathcal{D}) - \mathcal{L}(\theta_{t}; \mathcal{D}) \\
    \approx &
    \nabla_\theta^{\top} \mathcal{L}(\theta_{t}; \mathcal{D}) \cdot
    (\theta_{t+1} - \theta_{t}) \\
    := &
    \sum_{i=1}^K {{A_{lca}[t][i]}},
\label{eq:moment-lca}
\end{split}
\end{equation}
where each
$A_{lca}[t][i] = \nabla_{\theta^i} \mathcal{L}(\theta_t; \mathcal{D}) \cdot (\theta_{t+1}^i - \theta_{t}^i)$ and $A_{lca}[t][i]$ denotes LCA bound with the parameter $\theta_i$ at the update $t$.
Therefore, the loss change on $\mathcal{D}$ from update $t_1$
to update $t_2$ can be approximated by summing $(t_2 - t_1)$ equations
like Equation~\ref{eq:moment-lca} for all t in between:
\begin{equation}
\begin{split}
    & \mathcal{L}(\theta_{t_2}; \mathcal{D}) - \mathcal{L}(\theta_{t_1}; \mathcal{D}) \\
    \approx &
    \sum_{t=t_1}^{t_2 - 1}
    \mathcal{L}(\theta_{t + 1}; \mathcal{D}) - \mathcal{L}(\theta_{t}; \mathcal{D}) \\
    = & \sum_{t=t_1}^{t_2 - 1} \sum_{i=1}^K A_{lca}[t][i]
    = \sum_{i=1}^K {{\sum_{t=t_1}^{t_2 - 1} A_{lca}[t][i]}},
\end{split}
\label{eq:interval-lca}
\end{equation}
The above equation
denotes the so-called \textit{path integral}
of the loss change for each special
parameter $\theta^i$ along certain
optimization \textit{trajectory}
$\theta^i_{t_1}, \dots, \theta^i_{t_2}$, that is
$\sum_{t=t_1}^{t_2 - 1} A_{lca}[t][i]$.
This summation of \textit{moment} LCA values reflects the 
{\em effectiveness}
of $\theta_i$ with respect to the loss degradation
on certain dataset $\mathcal{D}$ between the update interval, 
so we call it the
\textit{interval} LCA value.

\subsection{Approximate LCA}
\label{sec:practical-consideration}

Theoretically,
the calculation of $\mathcal{L}(\theta_t, \mathcal{D})$ in
Equation~\ref{eq:moment-lca} requires forward computation of
the model over the whole dataset $\mathcal{D}$, which
will bring about too much computation overhead.
Instead, for each computation at update $t$, we only re-sample
a new mini-batch to be a representative of the whole dataset,
due to the previous smoothing trick,
we are actually evaluating a bootstrapping
of 15 mini-batches to represent the whole dataset,
which can reduce the variance to some extent.
In Section~\ref{sec:eval-approx},
we will empirically validate the rationality of this sampling approach with simulated experiments,
as an open question proposed in the original
LCA paper~\citep{lan-2019-loss-change-allocation}.

\subsection{Implementation Tricks}
Since the LCA values at each update $t$
should be stored to hard disk for subsequent
analyses, which represent the finest granuality
of the discrete learning dynamics,
this, in our investigation, will cause
large storage overheads with a model up
to millions trainable weights
and trained up to 100$K$ updates.
Therefore, in practice, we adopt two tricks in our implementation. 
Firstly, we store the LCA value once for every
15 updates via averaging the LCA values for those steps:
\begin{equation}
    \bar{A}_{lca}[t][i] = \frac{1}{15} \cdot \sum_{t'=15k + 1}^{15(k+1)} A_{lcs}[t'][i],
\end{equation}
for $k$ beginning with 0.
Then,
we divide the model parameters into several groups
and calculate LCA value for each group $g$
rather than each parameter $i\in g$ as follows:
\begin{equation}
    A_{lca}[t][g] = \frac{\sum_i \bar{A}_{lca}[t][i]}{|g|},
\end{equation}
where $\vert g \vert$ denotes the number of parameters that group. 
More precisely, we mainly study LCA for the following
parameter groups: word embedding in encoder (en\_emb);
$l$-th layer parameters in encoder (${\textrm{en}l}$);
$l$-th layer parameters in decoder (${\textrm{de}l}$);
word embedding in decoder (de\_emb);
softmax matrix in decoder (de\_softmax).



\section{Experiments and Analyses}

\subsection{Data and model}
We conduct experiments on two widely-used
translation benchmarks, namely IWSLT14 De$\Rightarrow$En  
and 
WMT14 En$\Rightarrow$De. 
We use the
Transformer base model~\citep{vaswani2017attention} from $\text{Fairseq}$~\citep{Ott2019fairseqAF}
for training and gathering the smoothed moment LCA values
of each model parameters. Thanks to the sampling technique in Section~\ref{sec:practical-consideration},
our training time is only doubled compared with standard training.
Our NMT system respectively achieves BLEU points of 34.4 and 27.7 on the test sets for IWSLT and WMT tasks, which are close to state-of-the-art.


\subsection{Evaluating the sampling approximation}
\label{sec:eval-approx}

\begin{figure}[t]
    \centering
    \includegraphics[scale=0.38]{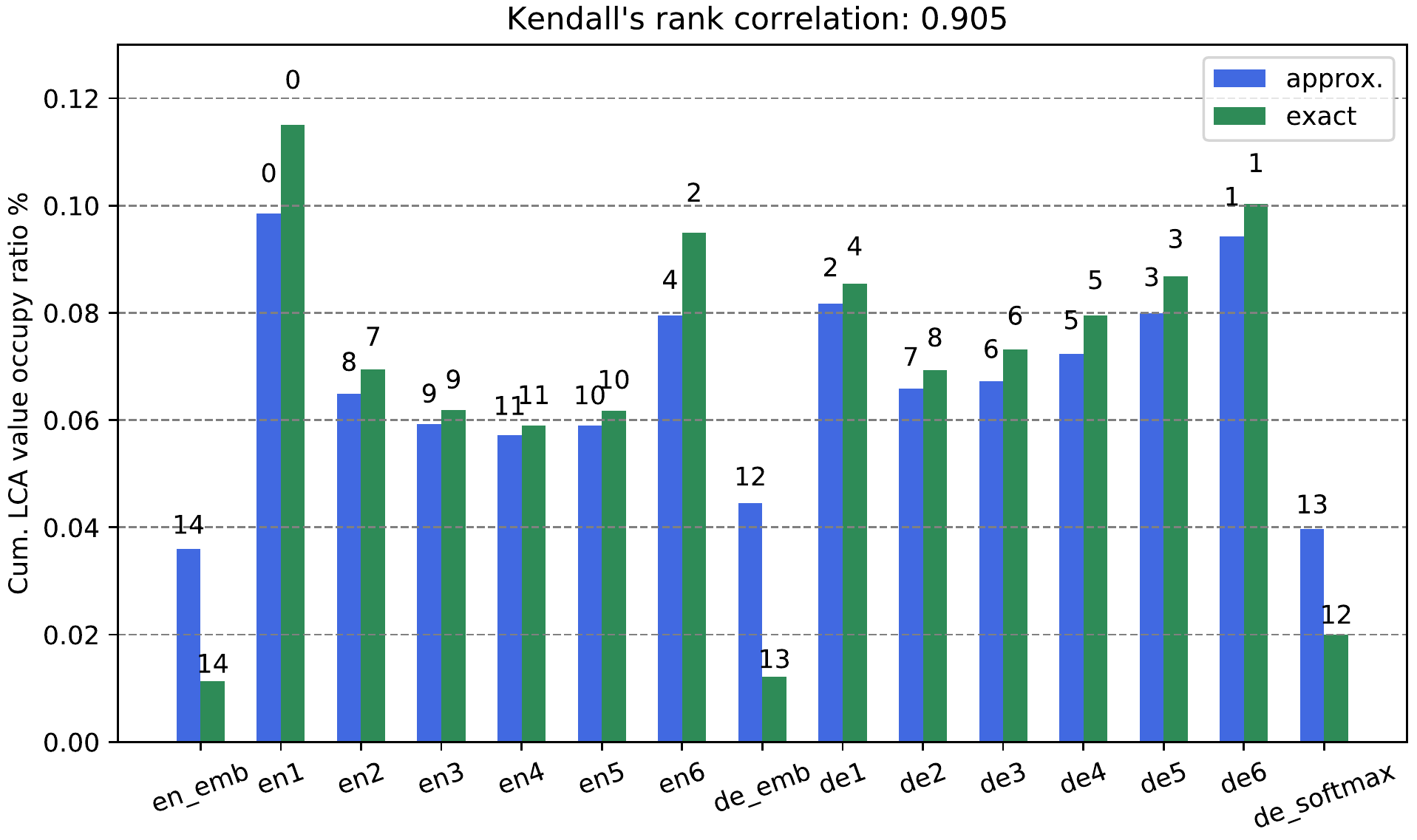}
    \caption{Evaluation of the sampling trick.
    }
    \label{fig:validate-sampling-trick}
\end{figure}

To prove the effectiveness of our approximation, we conduct a simulated experiment as follows: we randomly sample 10$K$ sentences from the IWSLT task and employ this small sampled data as the training data for running the exact implementation. 
Figure~\ref{fig:validate-sampling-trick}
demostrates cumulative LCA values' \textit{occupation ratio}s
of each module group in the Transformer described in \S2.3.
The ranking of each module's occupation ratio reflects
its \textit{relative} effectiveness
with respect to loss minimization on $\mathcal{D}$. So if the ranking of different
modules are similar between the sampling-based computation and exact computation,
we could relay on the sampling method for subsequent analyses.
As shown in Figure~\ref{fig:validate-sampling-trick}, the ranking similarity
between the sampling-based (approx.) and exact method are very similar to
each other, with a Kendall's rank coefficient as 0.905~\citep{kendall1938new}.



\subsection{Experimental analyses}

We conduct two main categories of analyses according to
LCA: i) \textit{interval} analysis: which tracks
the LCA values of each groups of model parameters
during certain interval $(t_1, t_2)$ of the whole training process;
ii) \textit{cumulative} analysis: which tracks the cumulative
LCA values from the beginning of training to the end. 

\subsubsection{Learning of sparse and dense weights}

\begin{figure}[t]
    \centering
    \includegraphics[scale=0.29]{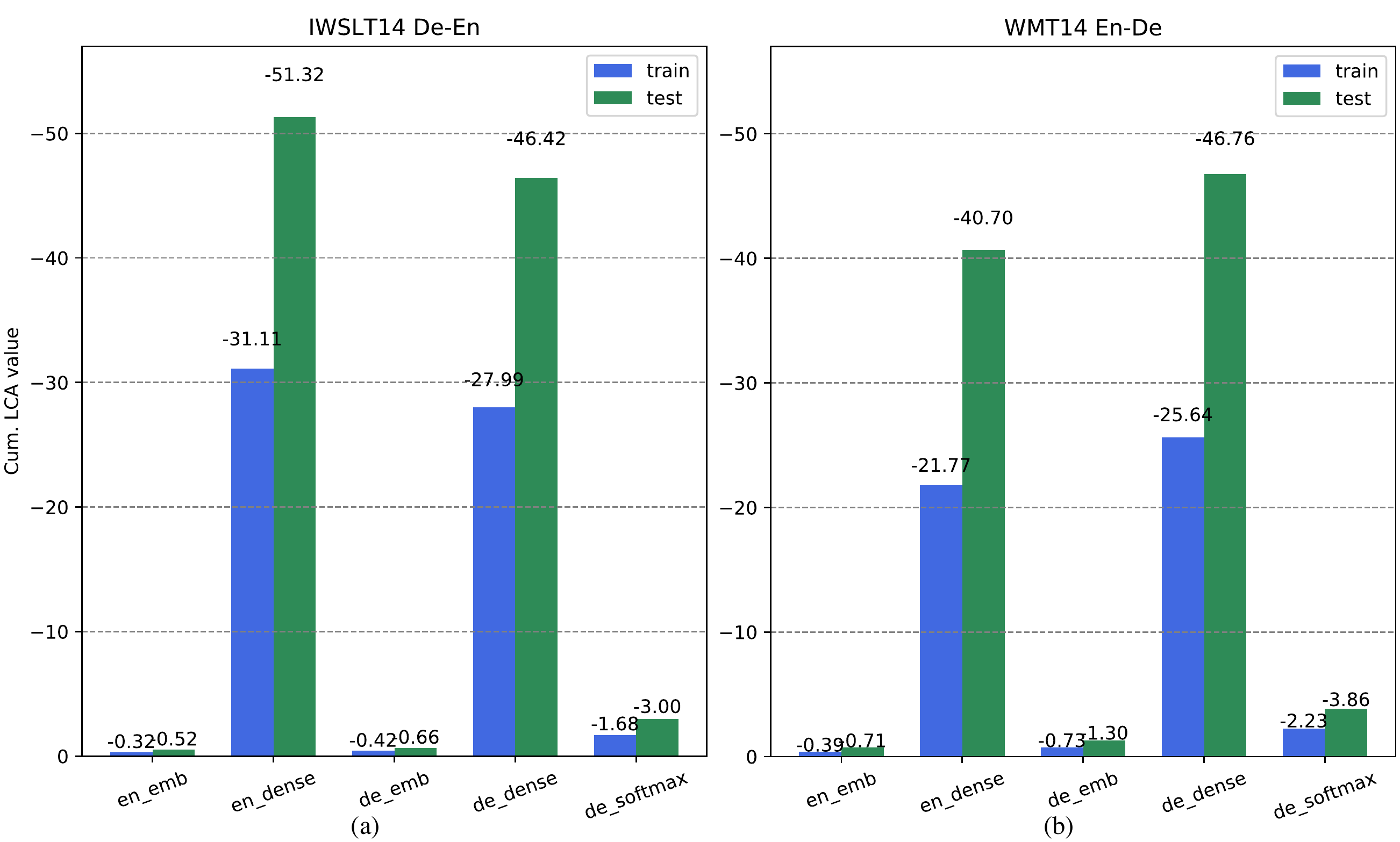}
    \caption{The cumulative LCA value of the sparse and dense weights of the Transformer.
    }
    \label{fig:enc-dec-modular-final-cum-lca}
\end{figure}


Current best practice sequence-to-sequence learning paradigm
proposes an explicit differentiation between encoder
and decoder. This explicit separation \textit{may} provide a bottleneck of
gradient flow from the loss to the encoder. We visualize the
cumulative LCA value of sparse and dense weights of Transformer
in Figure~\ref{fig:enc-dec-modular-final-cum-lca}.

Overall speaking,
dense weights from encoder and decoder contribute
similarly both on train and test. However, the sparse
embeddings both contribute very little.
This might because that the unpdate frequency of
dense weights is much larger than the sparse weights.
However, the dense softmax weigths's LCA value
(decoder's output embedding)  is still far less than those
middle layers.

\subsubsection{Layer-wise learning dynamics}
\label{sec:layer-wise-learndyn}

\begin{figure*}[!ht]
    \centering
    \includegraphics[scale=0.45]{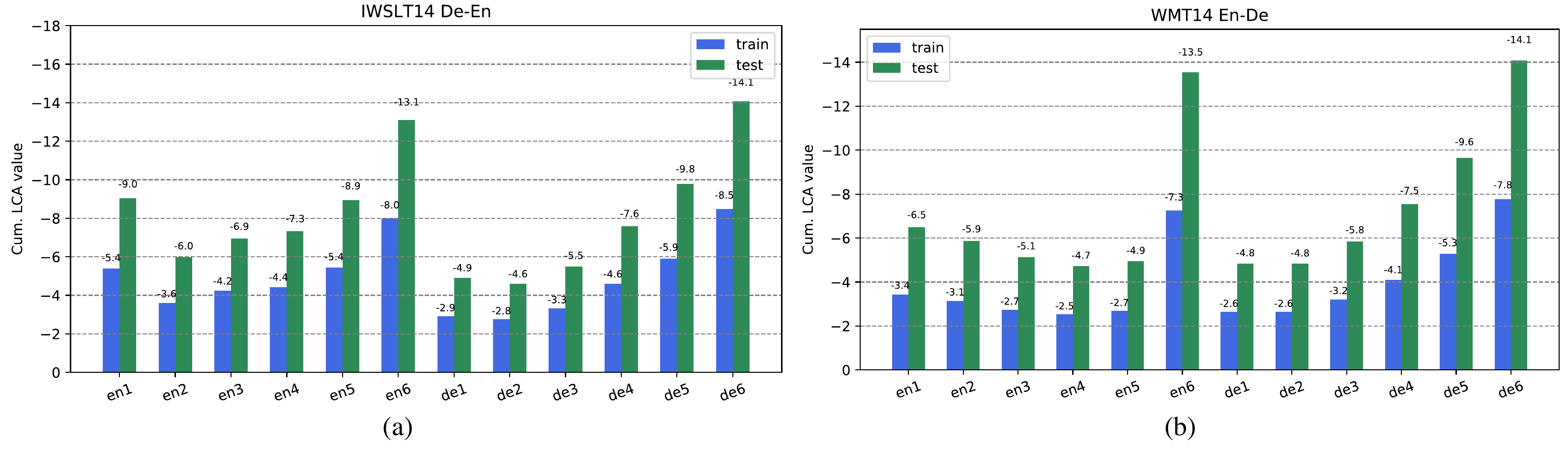}
    \caption{The cumulative LCA value of the layerwise dense parameters of the encoder and decoder on the two datasets, as a zoom-in of Figure~\ref{fig:enc-dec-modular-final-cum-lca}'s \textit{en/de\_dense} bars.}
    \label{fig:layerwise-final-cum-lca}
\end{figure*}

\begin{figure*}[t]
    \centering
    \includegraphics[scale=0.49]{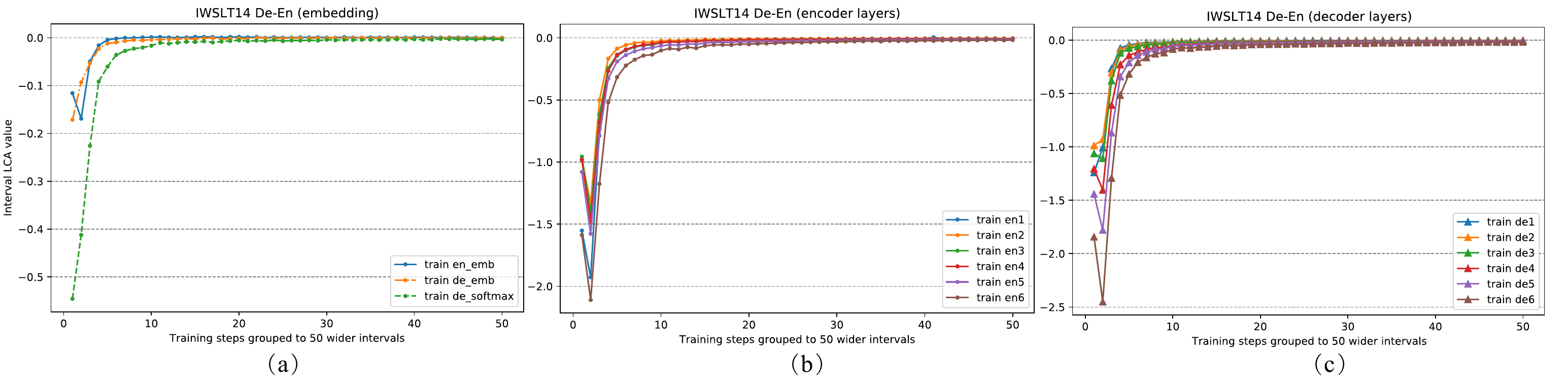}
    \caption{The interval LCA value on the IWSLT14 dataset allocated on different modules along training.
    }
    \label{fig:lca-interval-iw14}
\end{figure*}

To further analyze the layer-wise contribution of each
encoder and decoder layer, we summarize the
cumulative LCA value of each dense layer in Figure~\ref{fig:layerwise-final-cum-lca}.
There is an interesting sandwitch effect of the encoder where the beginning
and the end layer contribute the most while the layer in between contribute
less.
For the decoder layers, the more higher the layer, the more contribution
it makes to the loss change. It is very clear that from this modular view,
different neural blocks provide similar effect on loss change on train as on test.

To further understand the convergence property shown in \citet{NIPS2017_7188}:
that lower layer converges earlier. We draw the grouped interval LCA values along
the whole training process in Figure~\ref{fig:lca-interval-iw14}. As you can see,
higher layers tend to have smaller LCA values which means they contribute more
than lower layers generally at any training interval, functioning as an evidence that
higher layers continue to evolve representations.

\subsubsection{Learning of the embeddings}
\label{sec:emb}


As the vocabulary size is large, we can not visualize the behaviors for 
the embedding of every word in the vocabulary. We thereby split the vocabulary into 25 groups according to word frequency.

\begin{figure}[!h]
    \centering
    \includegraphics[scale=0.28]{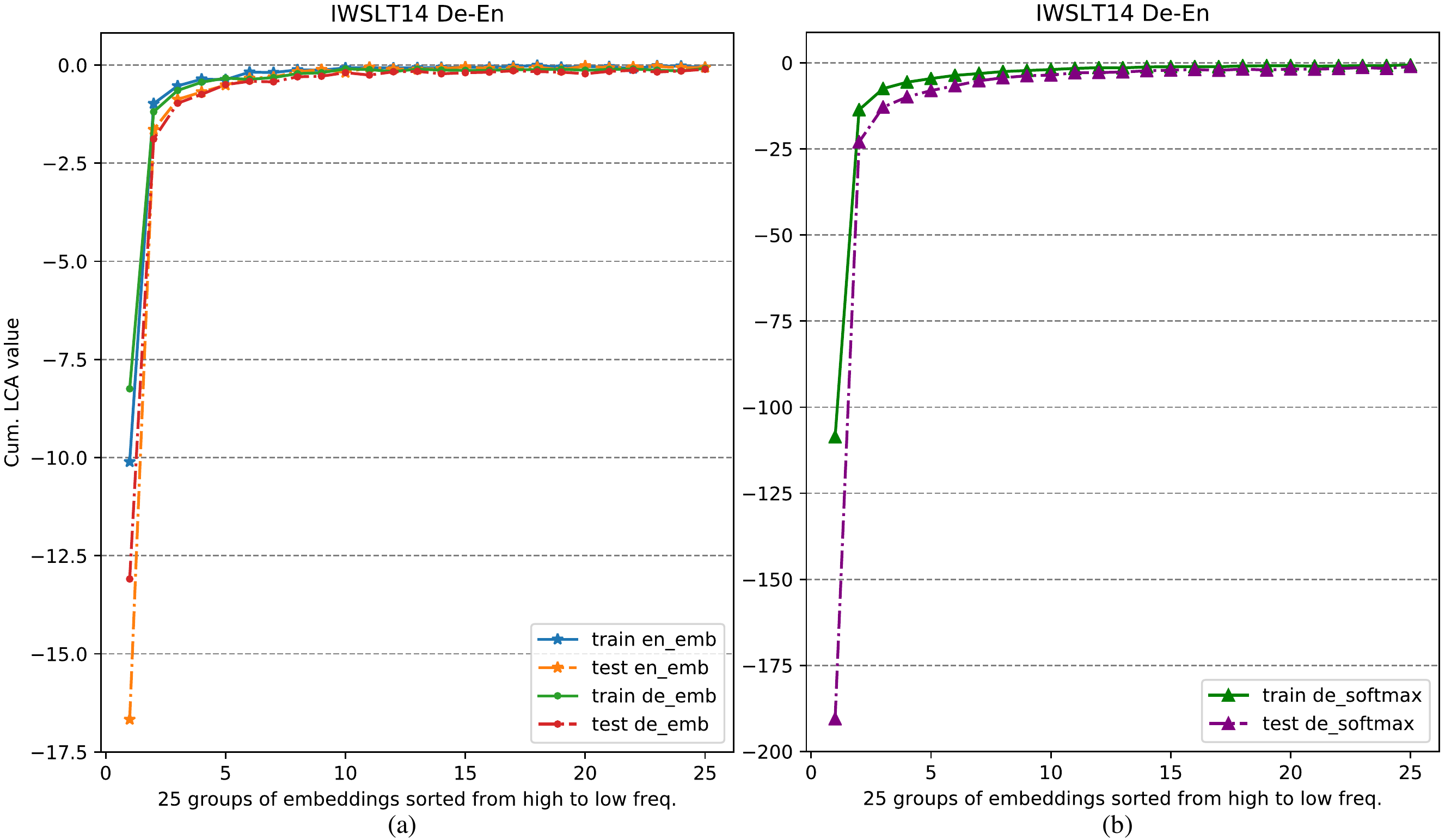}
    \caption{The cumulative final LCA value of sparse and dense embeddings divided by word frequency groups.}
    \label{fig:emb-vec-final-lca-iw14}
\end{figure}

\noindent
Figure~\ref{fig:emb-vec-final-lca-iw14} visualizes the
cumulative LCA values for all groups
sorted by word frequency.
From this Figure, one can clearly see that words with very
high frequency occupy most LCA values than lower frequency words on both training and test datasets. This fact further provides an explanation for the well-known question, i.e. why infrequent words are difficult to be translated but frequent words are easy for NMT.

\section{Conclusion}

In this paper we propose to use Loss Change Allocation (LCA)~\citep{lan-2019-loss-change-allocation} for understanding learning dynamics of NMT. Since the exact calculation of LCA requires calculating the gradient on an entire dataset at each update, we instead present an approximate to put it into practice in NMT scenario.
Our simulated experiment shows that such approximate calculation is efficient and is empirically proved to deliver consistent results to the exact implementation. Extensive experiments on two standard translation tasks reveal some valuable findings: parameters of encoder (decoder) word embeddings and softmax matrix contribute less to loss decrease and those of the first layer in encoder
and the last layer in decoder contribute most to loss decrease during the training process.
We will investigate the the relation of loss decrease with other interesting learning phenomenon,
for example the emergence of weight sparsity~\citep{voita-etal-2019-analyzing,NIPS2019_9551} and module criticality~\citep{zhang2019all,Chatterji2019TheIR}.

\bibliography{acl2019}
\bibliographystyle{acl_natbib}

\end{document}